# Not As Easy As It Seems: Automating the Construction of Lexical Chains Using *Roget's Thesaurus*

Mario Jarmasz and Stan Szpakowicz

School of Information Technology and Engineering
University of Ottawa
Ottawa, Canada, K1N 6N5
{mjarmasz, szpak}@site.uottawa.ca

**Abstract.** Morris and Hirst [10] present a method of linking significant words that are about the same topic. The resulting *lexical chains* are a means of identifying cohesive regions in a text, with applications in many natural language processing tasks, including text summarization. The first lexical chains were constructed manually using *Roget's International Thesaurus*. Morris and Hirst wrote that automation would be straightforward given an electronic thesaurus. All applications so far have used *WordNet* to produce lexical chains, perhaps because adequate electronic versions of *Roget's* were not available until recently. We discuss the building of lexical chains using an electronic version of *Roget's Thesaurus*. We implement a variant of the original algorithm, and explain the necessary design decisions. We include a comparison with other implementations.

## 1 Introduction

Lexical chains [10] are sequences of words in a text that represent the same topic. The concept has been inspired by the notion of cohesion in discourse [7]. A sufficiently rich and subtle lexical resource is required to decide on semantic proximity of words.

Computational linguists have used lexical chains in a variety of tasks, from text segmentation [10], [11], to summarization [1], [2], [12], detection of malapropisms [7], the building of hypertext links within and between texts [5], analysis of the structure of texts to compute their similarity [3], and even a form of word sense disambiguation [1], [11]. Most of the systems have used *WordNet* [4] to build lexical chains, perhaps in part because it is readily available. An adequate machine-tractable version of *Roget's Thesaurus* has not been ready for use until recently [8]. The lexical chain construction process is computationally expensive but the price seems worth paying if we then can incorporate lexical semantics in natural language systems.

We build lexical chains using a computerized version of the 1987 edition of Penguin's *Roget's Thesaurus of English Words and Phrases* [8], [9]. The original lexical chain algorithm [10] exploits certain organizational properties of *Roget's*. *WordNet*-based implementations cannot take advantage of *Roget's* relations. They also usually only link nouns, as relations between parts-of-speech are limited in *WordNet*. Morris and Hirst wrote: "Given a copy [of a machine readable thesaurus], implementation [of lexical chains] would clearly be straightforward". We have set out to test this statement in practice. We present a step-by-step example and compare existing methods of evaluating lexical chains.

## 2 Lexical Chain Building Algorithms

Algorithms that build lexical chains consider one by one the words for inclusion in the chains constructed so far. Important parameters to consider are the lexical resource used, which determines the lexicon and the possible thesaural relations, the thesaural relations themselves, the transitivity of word relations and the distance — measured in sentences — allowed between words in a chain [10].

Barzilay and Elhadad [2] present the following three steps:

1. Select a set of candidate words;
2. For each candidate word, find an appropriate chain relying on a relatedness criterion among members of the chain;
3. If it is found, insert the word in the chain and update it accordingly.

*Step1: Select a set of candidate words.* Repeated occurrences of closed-class words and high frequency words are not considered [10]. We remove words that should not appear in lexical chains, using a 980-element stop list, union of five publicly-available lists: Oracle 8 ConText, SMART, Hyperwave, and lists from the University of Kansas and Ohio State University. After eliminating these high frequency words it would be beneficial to identify nominal compounds and proper nouns but our current system does yet not do so. *Roget's* allows us to build lexical chains using nouns, adjectives, verb, adverbs and interjections; we have therefore not found it necessary to identify the part-of-speech. Nominal compounds can be crucial in building correct lexical chains, as argued by [1]; considering the words *crystal* and *ball* independently is not at all the same thing as considering the phrase *crystal ball*. *Roget's* has a very large number of phrases, but we do not take advantage of this, as we do not have a way of tagging phrases in a text. There are few proper nouns in the *Thesaurus*, so their participation in chains is limited.

*Step 2: For each candidate word, find an appropriate chain.* Morris and Hirst identify five types of thesaural relations that suggest the inclusion of a candidate word in a chain [10]. We have decided to adopt only the first one, as it is the most frequent relation, can be computed rapidly and consists of a large set of closely related words. We also have simple term repetition. The two relations we use, in terms of the 1987 *Roget's* structure [8], are:

1. Repetition of the same word, for example: *Rome*, *Rome*.
2. Inclusion in the same *Head*. *Roget's Thesaurus* is organized in 990 Heads that represent concepts [8], for example: **343** *Ocean*, **747** *Restraint* and **986** *Clergy*. Two words that belong in the same head are about the same concept, for example: *bank* and *slope* in the Head **209** *Height*.

A Head is divided into paragraphs grouped by part-of-speech: nouns, adjectives, verbs and adverbs. A paragraph is divided into semicolon groups of closely related words, similar to a *WordNet* synset, for example *{mother, grandmother 169 maternity}* [8]. There are four levels of semantic similarity within a Head: two words or phrases located in the same semicolon group, paragraph, part-of-speech and Head.

Morphological processing must be automated to assess the relation between words. This is done both by *WordNet* and the electronic version of *Roget's*. Relations between words of different parts-of-speech seem to create very non-intuitive chains, for example: *{constant, train, train, rigid, train, takes, line, takes, train, train}*. The adjective *constant* is related to *train* under the Head **71** *Continuity: uninterrupted sequence* and *rigid* to *train* under the Head **83** *Conformity*, but these words do not seem to make sense in the context of this chain. This relation may be too broad when applied to all parts-of-speech. We have therefore decided to restrict it to nouns. *Roget's* contains around 100 000 words [8], but very few of them are technical. Any word or phrase that is not in the *Thesaurus* cannot be linked to any other except via simple repetition.

*Step 3: Insert the word in the chain.* Inclusion requires a relation between the candidate word and the lexical chain. This is the essential step, most open to interpretation. An example of a chain is *{cow, sheep, wool, scarf, boots, hat, snow}* [10]. Should all of the words in the chain be close to one another? This would mean that *cow* and *snow* should not appear in the same chain. Should only specific senses of a word be included in a chain? Should a chain be built on an entire text, or only segments of it? Barzilay [1] performs word sense disambiguation as well segmentation before building lexical chains. In theory, chains should disambiguate individual senses of words and segment the text in which they are found; in practice this is difficult to achieve. What should be the distance between two words in a chain? These issues are discussed by [10] but not definitively answered by any implementation. These are serious considerations, as it easy to generate spurious chains. We have decided that all words in a chain should be related via a thesaural relation. This allows building cohesive chains. The text is not segmented and we stop building a chain if no words have been added after seeing five sentences.

*Step 4: Merge lexical chains and keep the strongest ones.* This step is not explicitly mentioned by Barzilay [1] but all implementations perform it at some point. The merging algorithm depends on the intermediary chains built by a system. Section 4 discusses the evaluation of the strength of a chain.

## 3 Step-by-Step Example of Lexical Chain Construction

Ellman [4] has analyzed the following quotation, attributed to Einstein, for the purpose of building lexical chains. The words in bold are the candidate words retained by our system after applying the stop list.

*We* ***suppose*** *a very long* ***train travelling*** *along the* ***rails*** *with a* ***constant velocity*** *v and in the* ***direction*** *indicated in* ***Figure*** *1.* ***People travelling*** *in this* ***train*** *will with* ***advantage*** *use the* ***train*** *as a* ***rigid reference-body****; they* ***regard*** *all* ***events*** *in* ***reference*** *to the* ***train****. Then every* ***event*** *which* ***takes*** *place along the* ***line*** *also* ***takes*** *place at a particular point of the* ***train****. Also, the* ***definition*** *of* ***simultaneity*** *can be given* ***relative*** *to the* ***train*** *in exactly the same way as with* ***respect*** *to the* ***embankment****.*

All possible lexical chains (consisting of at least two words) are built for each candidate word, proceeding forward through the text. Some words have multiple chains, for example *{direction, travelling, train, train, train, line, train, train}*, *{direction, advantage, line}* and *{direction, embankment}*. The strongest chains are selected for each candidate word. A candidate generates its own set of chains, for example *{events, train, line, train, train}* and *{takes, takes, train, train}*. These two chains can be merged if we allow one degree of transitivity: *events* is related to *takes* since both are related to *train*. Once we have eliminated and merged chains, we get:

1. *{train, travelling, rails, velocity, direction, travelling, train, train, events, train, takes, line, takes, train, train, embankment}*
2. *{advantage, events, event}*
3. *{regard, reference, line, relative, respect}*

As a reference, the chains can be compared to the eight obtained by Ellman [4]: 1. *{train, rails, train, line, train, train, embankment}*, 2. *{direction, people, direction}*, 3. *{reference, regard, relative-to, respect}*, 4. *{travelling, velocity, travelling, rigid}*, 5. *{suppose, reference-to, place, place}*, 6. *{advantage, events, event}*, 7. *{long, constant}*, 8. *{figure, body}*. There also are nine chains obtained by St-Onge [4]: 1. *{train, velocity, direction, train, train, train, advantage, reference, reference-to, train, train, respect-to, simultaneity}*, 2. *{travelling, travelling}*, 3. *{rails, line}*, 4. *{constant, given}*, 5. *{figure, people, body}*, 6. *{regard, particular, point}*, 7. *{events, event, place, place}*, 8. *{definition}*, 9. *{embankment}*. We do not generate as many chains as Ellman or St-Onge, but we feel that our chains adequately represent the paragraph. Now we need an objective way of evaluating lexical chains.

## 4 Evaluating Lexical Chains

Two criteria govern the evaluation of a lexical chain: its strength and its quality. Morris and Hirst [10] identified three factors for evaluating strength: reiteration, density and length. The more repetitious, denser and longer the chain, the stronger it is. This notion has been generally accepted, with the addition of taking into account the type of relations used in the chain when scoring its strength [2], [3], [8], [12].

There should be an objective evaluation of the quality of lexical chains, but none has been developed so far. Existing techniques include assessing whether a chain is intuitively correct [4], [10]. Another technique involves measuring the success of lexical chains in performing a specific task, for example the detection of malapropisms [8], text summarization [2], [3], [12], or word sense disambiguation [1], [11]. Detection of malapropisms can be measured using precision and recall, but a large annotated corpus is not available. The success at correctly disambiguating word senses can also be measured, but requires a way of judging if this has been done correctly. [1] relied on a corpus tagged with *WordNet* senses, [11] used human judgment. There are no definite ways of evaluating text summarization.

## 5 Discussion and Future Work

We have shown that it is possible to create lexical chains using an electronic version of *Roget's Thesaurus*, but that it is not as straightforward as it originally seemed. *Roget's* has a much richer structure for lexical chain construction than exploited by [10]. Their thesaural relations are too broad to build well-focused chains or too computationally expensive to be of interest. *WordNet* implementations have different sets of relations and scoring techniques to build and select chains. Although there is a consensus on the high-level algorithm, there are significant differences in implementations. The major criticism of lexical chains is that there is no adequate evaluation of their quality. Until it is established, it will be hard to compare implementations of lexical chain construction algorithms. We plan to build a harness for testing the various parameters of lexical chain construction listed in this paper. We expect to propose a new evaluation procedure. For the time being, we intend to use a corpus containing tagged malapropisms.

## Acknowledgments

We thank Terry Copeck for having prepared the stop list used in building the lexical chains. This research would not have been possible without the help of Pearson Education, the owners of the 1987 Penguin's *Roget's Thesaurus of English Words and Phrases*. Partial funding for this work comes from NSERC.